\documentclass[letter]{article}

\usepackage{spconf,amsmath,graphicx}
\usepackage{times}
\usepackage{latexsym}
\usepackage{amsmath}
\usepackage{url}
\usepackage{subcaption}
\usepackage{graphicx}

\usepackage{caption}
\usepackage{multirow}
\usepackage{amssymb}
\usepackage{mathrsfs}
\usepackage{graphics}
\usepackage{graphicx}
\usepackage{ifsym}
\usepackage{amsmath}
\usepackage{subcaption}
\usepackage{booktabs}
\usepackage{multirow}
\usepackage{array}
\usepackage{bbm}
\usepackage{multicol}
\usepackage{blindtext}
\usepackage{algorithm}
\usepackage{algpseudocode}
\usepackage{tikz}

\usepackage{lipsum}
\usepackage{ntheorem}
\usepackage{makecell}

\let\oldbibliography\thebibliography
\renewcommand{\thebibliography}[1]{%
  \oldbibliography{#1}%
  \setlength{\itemsep}{0pt}%
}

\DeclareMathOperator*{\argmax}{arg\,max}
\DeclareMathOperator*{\argmin}{arg\,min}

\DeclareCaptionFont{10pt}{\fontsize{10pt}{12pt}\selectfont}
\title{\textbf{An Adversarial Learning based Multi-Step Spoken Language Understanding System through Human-Computer Interaction}}
\name{Yu Wang, Yilin Shen and Hongxia Jin}
\address{AI Center, Samsung Research America, Mountain View, CA, USA}

\begin{document}

\maketitle
\begin{abstract}
Most of the existing spoken language understanding systems can perform only semantic frame parsing based on a single-round user query. They cannot take users' feedback to update/add/remove slot values through multiround interactions with users. In this paper, we introduce a novel multi-step spoken language understanding system based on adversarial learning that can leverage the multiround user's feedback to update slot values. We perform two experiments on the benchmark ATIS dataset and demonstrate that the new system can improve parsing performance by at least $2.5\%$ in terms of F1, with only one round of feedback. The improvement becomes even larger when the number of feedback rounds increases. Furthermore, we also compare the new system with state-of-the-art dialogue state tracking systems and demonstrate that the new interactive system can perform better on multiround spoken language understanding tasks in terms of slot- and sentence-level accuracy.
\end{abstract}

\section{Introduction}
Semantic frame parsing is an important research topic in spoken language understanding (SLU). The main target of semantic frame parsing in SLU is to extract meaningful slots from the query and assign them correct slot tags, \emph{i.e.,} slot filling. Traditionally, semantic frame parsing can be achieved by a variety of techniques, including conditional random fields (CRFs) \cite{xu2013convolutional,yao2014recurrent}, hidden Markov chains (HMMs) \cite{he2003hidden} and support vector machines (SVMs) \cite{raymond2007generative}. Recent works on semantic frame parsing have sought to leverage recurrent neural network (RNN) models for sequence prediction \cite{xu2013convolutional,liu2016attention,wang2018new,wang2018deep,goo2018slot,wang2018bi,wang2020multi,chen2019bert,zhang2018joint,wang2020new}. 

Many models demonstrate decent performance on different benchmark SLU datasets, such as ATIS \cite{price1990evaluation} and SNIPS \cite{coucke2018snips}. Most of these models are only able to perform semantic frame parsing based on single-round question answering (QA) between a user and a machine. The system cannot correct the slot labels based on a user's feedback if its first round response is wrong, let alone take extra information from a user's response if his/her first-round query is incomplete.

Currently, people use dialogue state tracking (DST) models to handle multiturn responses in a dialogue system \cite{kumar2020ma,wu2019transferable,wang2019deep,wang2020interactive}. These models, however, mostly focus on handling topic changing, multidomain adaption and improving goal accuracy, which are very different from the target of multiround semantic frame parsing. The main differences between multiround semantic frame parsing and DST are the following:\\
1. Normally, there is no topic change in a multiround semantic frame parsing task in comparison to that in a DST task.\\
2. The system's feedback is not necessarily the text, as it is in a DST task. More importantly, it is common that the system's feedback is not part of our training data since we only care about the user's responses most of the time.\\
3. Normally, there is no domain adaptation in a multiround semantic frame parsing task in comparison to in a DST task.

To handle these multiround scenarios by understanding continuous feedback from users, in this paper, we propose a human-computer Multi-Step SLU (MS$_2$LU) model, which can learn a robust reward function through human demonstration using inverse reinforcement learning (IRL), such that multiround semantic frame parsing (or slot filling) can be achieved.

\begin{figure}[tph!]
\centering
\includegraphics[width=0.52\textwidth,height=0.34\textwidth]{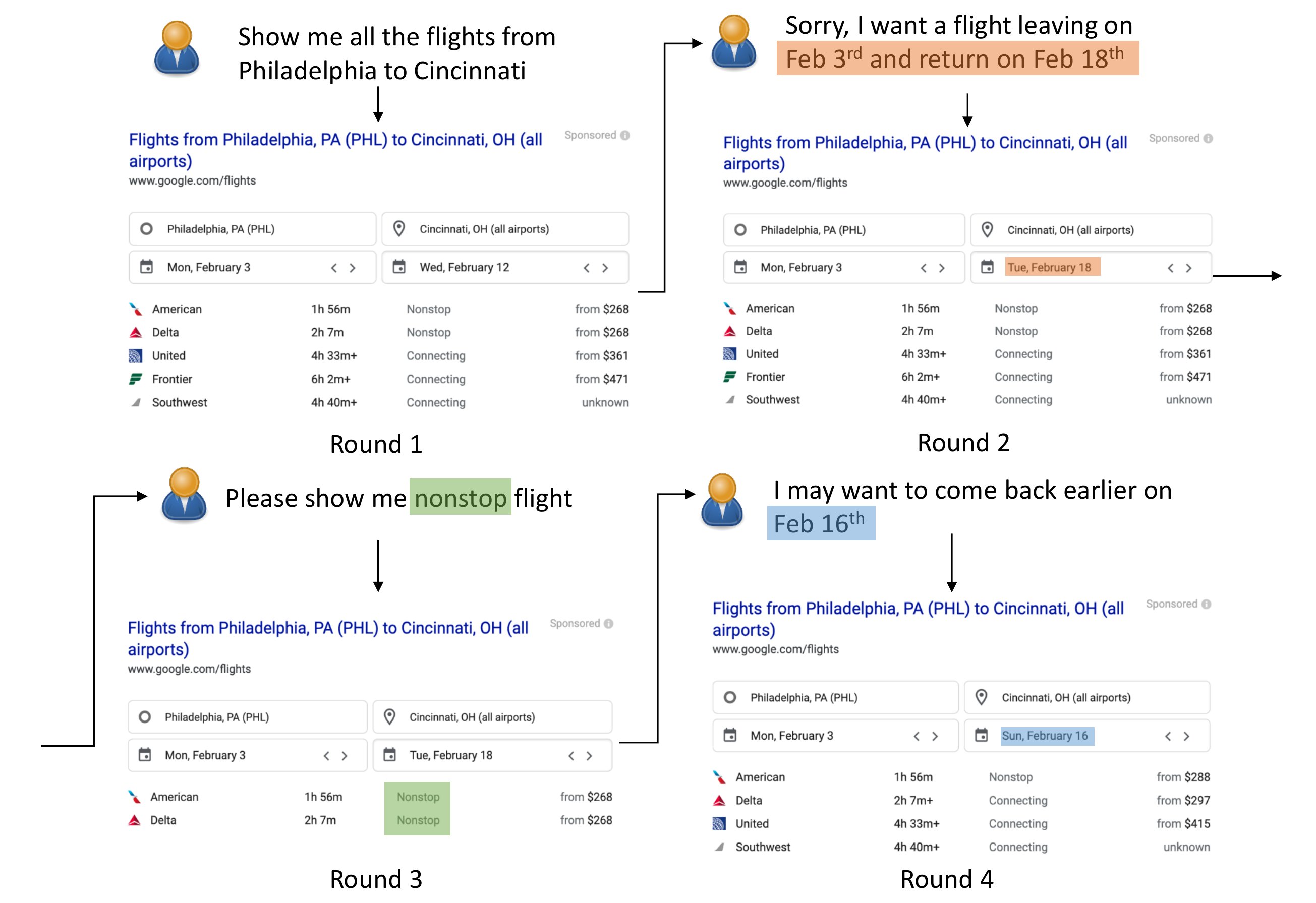}
\caption{A flight booking example using the MS$_2$LU model on the ATIS dataset}
\label{fig:fig1}
\end{figure}

Figure \ref{fig:fig1} demonstrates a flight booking example generated from the ATIS dataset by using our MS$_2$LU model. The result at each step is generated by sending our extracted slot information to the flight booking API. During his/her interaction with our system, a user provides several pieces of additional information by adding a leaving/returning date and flight type in Rounds 2 and 3, respectively. In Round 4, the user even changes his/her returning date again to another date. It can be observed that our system can handle all these changes together with new information robustly, update and extract the corresponding slot tags accurately, and finally fetch the results from flight API correctly. All these features benefit from the MS$_2$LU system introduced in this paper.

The contributions of this work are threefold:\\
1. We propose a novel semantic frame parsing framework using a multi-step learning technique to achieve multiround slot filling based on user demonstration and feedback.\\
2. We evaluate our model on the benchmark ATIS SLU dataset. Our system generates real flight information retrieved from Google Flight API and allows for one round of user feedback to correct if the answer is wrong during training and testing. This achieves the state-of-the-art performance on the test dataset.\\
3. We design and perform a multiround human flight booking task via Amazon Mechanical Turk based on the ATIS dataset to demonstrate the robustness of the model for handling extra slot information and updating the slot values during runtime.

The paper is organized as follows:
Section 2 presents all the details about the multi-step SLU framework using reinforcement learning \cite{wang2018boosting,wang2018new}and adversarial learning. The system contains several components---the feature generator, the slot extraction model, the adversarial discriminator and the reward estimator---to generate rewards using IRL. In Section 3, we conduct two experiments on the ATIS dataset: one is an SLU task with one round of user feedback, and the other is a multiround SLU flight booking task via Amazon Mechnical Turk.
\section{Multi-Step SLU (MS$_2$LU) System}
\begin{figure*}[tph!]
\centering
\includegraphics[width=0.8\textwidth,height=0.42\textwidth]{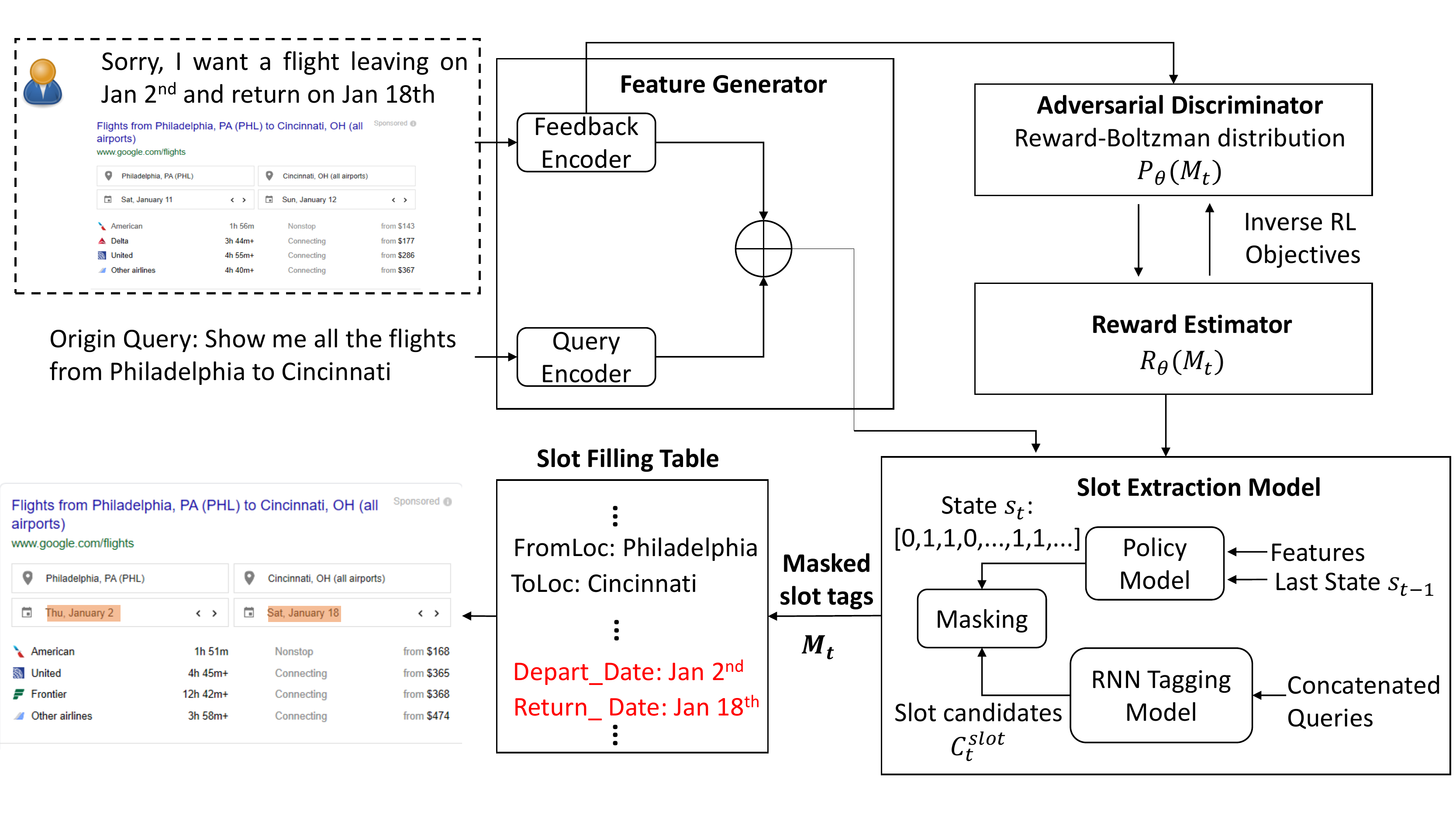}
\caption{Overview of the Multi-Step SLU System}
\label{fig:fig2}
\end{figure*}

The MS$_2$LU system is a reinforcement learning-based semantic frame parsing framework that can leverage the user's response step by step to improve its system performance. Figure \ref{fig:fig2} shows a detailed graphical description of our MS$_2$LU system structure. The system contains four main submodules: the feature generator, the slot extraction model, the reward estimator and the adversarial discriminator. Their designs are detailed as follows.
\subsection{Feature Generator}
The feature generator extracts important semantic features from the origin query and user feedback at each round. These features are used to estimate reward $R$ and to generate policy $\pi$ and state $s_t$ in the slot extraction model.

Specifically, both the origin query and user feedback are encoded by the attention bidirectional RNN structure, as given in \cite{bahdanau2014neural}. The encoded query feature is denoted as $c^{q}$, which is the final output of the query encoder $E_{query}$. Comparatively, the encoded user feedback feature $c^{f}_t$ is different at each time step (or round) $t$ because it only considers the user's feedback feature generated by the feedback encoder $E_{feedback}$ in round $t$. The two feature embeddings are then concatenated together as [$c^{q}$,$c^{f}_t$] to be consumed by the reward estimator and slot extraction model.
\subsection{Slot Extraction Model}
The slot extraction model contains two submodules: a reinforcement learning (RL) policy model and an RNN tagging model.

The RL policy model generates policy $\pi_\beta$ and the masking rules, \emph{i.e.,} state $s_t \in \mathbb{R}^{1\times k}$, to identify the slot values to be maintained. Each of $s_t$'s entry is a binary value of either 0 or 1, representing whether the $i^{th}$ label $l_i$ should be presented (1) or not (0). The state update can be represented mathematically as follows:
\begin{equation}
s_t = f_{\pi_\beta}(g[c^q,c_t^f],s_{t-1})
\end{equation}
where $f (\cdot)$ is the long short-term memory (LSTM) unit \cite{hochreiter1997long}, and $g(\cdot)$ is a multilayer perception (MLP) for origin query and user feedback matching. $\pi_\beta$ is the estimated RL policy with parameter $\beta$. The update law of $\beta$ is handled by the adversarial discriminator to be discussed.

The RNN tagging model extracts the slot candidate matrix $C^{slot}_t\in \mathbb{R}^{k\times m}$ from the concatenated queries $[c^q,c_t^f]$. The $i^{th}$ row of $C^{slot}_t$ represents the average sum of the labeled token embeddings under the $i^{th}$ label $l_i$, $k$ is the total number of label types, and $m$ is the word embedding dimension. If there is no token under a label, then that label's row is padded by zeros.

The final output of the slot extraction model is the masked slot matrix $M_t$ defined as follows:
\begin{equation}
M_t = diag(s_t) \cdot C^{slot}_t
\end{equation}
where $diag(s_t) \in \mathbb{R}^{k\times k}$ is a diagonal matrix with its diagonal element $diag(s_t)(i,i)=s_t(i)$, and all other elements are zeros. Therefore, $M_t\in \mathbb{R}^{k \times m}$ is the masked slot candidate matrix by only keeping the rows where $s_t$ has nonzero values (and leaving the other rows with zeros).
\subsection{Reward Estimator}
The reward estimator $R_\theta(M_t)$ is a nonlinear function $\phi{\cdot}$, which takes the current round masked slot matrix $M_t$ and user feedback feature vector $c^{f}_t$ as its input.

Inspired by the reward design in different IRL applications \cite{li2010contextual,hadfield2016cooperative,wang2018no,li2019dialogue}, the reward estimator function $R_\theta(M_t)$ is defined as follows:
\begin{equation}
R_\theta(M_t)=\phi(W_t(f(M_t)+M_tc^{f}_t)+b_t)
\end{equation}
where $\phi$ is a nonlinear projection function, and $W_t$ and $b_t$ denote the weight and bias in the output layer, respectively. $M_tc^f_t$ stands for the projection of the feedback feature $c^f_t$ to the slot filling representation space $M_t$, and $f(\cdot)$ is an LSTM structure. \\
{\it{Remarks:}} It is worth noting that both the slot candidate matrix $C_t^{slot}$ and the masked matrix $M_t$ depend only on the origin query and the current-round $t$'s user feedback. The state $s_t$ contains previous-round user feedback information and hence can help filter out the unnecessary slot values and only keep the useful information.
\subsection{Adversarial Discriminator}
To train the nonlinear estimated reward function $R_{\theta}(M_t)=\phi(\cdot)$, we leverage the adversarial discriminator to associate the generated masked slot candidate matrix with the reward function. Similar to \cite{wang2018no}, we use the reward Boltzman distribution to approximate the data distribution:
\begin{equation}
p_{\theta}(M_t)=\dfrac{e^{R_\theta(M_t)}}{\sum_{M_i}e^{R_\theta(M_i)}}
\end{equation}
where $M_i$ denotes the empirical sample at time step $i$. The optimal reward function $R^*(M)$ is achieved when the reward Boltzman distribution $p_{\theta}(M_t)$ is equal to that in the ``real" data distribution $p^*(M)$.

The objective function $J$ of the adversarial discriminator is a min-max function that maximizes $p_\theta$'s similarity with the empirical distribution of the training data $p_e$ and minimizes the similarity between them and the data generated by our slot extraction policy $\pi_{\beta}$, which can be mathematically represented as follows:
\begin{equation}
J=\argmax_{\beta} \argmin_{\theta}KL(p_e \mid\mid p_{\theta})-KL(\pi_{\beta} \mid\mid p_{\theta})
\end{equation}
where $KL(\cdot)$ represents the KL divergence.
Following a similar derivation in \cite{wang2018no}, our SGD learning law for the policy model network parameter $\beta$ and the reward estimator network parameter $\theta$ can be written as follows:
\begin{equation}
\begin{split}
\dfrac{\partial J_\theta}{\partial \theta} &= \mathbb{E}_{M\sim p_e(M)}(\dfrac{\partial R_\theta}{\partial \theta})-\mathbb{E}_{M\sim \pi_\beta(M)}(\dfrac{\partial R_\theta}{\partial \theta})\\
\dfrac{\partial J_\beta}{\partial \beta} &= \mathbb{E}_{M\sim \pi_\beta(M)}(R_\theta(M)-\log\pi_{\beta}(M)-b)\\&\;\;\;\;\times\dfrac{\log\pi_{\beta}(M)}{\partial \beta}
\end{split}
\end{equation}
\subsection{Slot Filling Table}
As shown in Figure \ref{fig:fig2}, the slot extraction model sends its output $M_t$ to the slot filling table $T$ to update the corresponding slot entries. The slot filling table at round $t$ is represented by $T_{t-1}$, containing $k$ rows, where $k$ is the number of slot types. At each round $t$, the table value $T_t$ is updated as follows:
\begin{equation}
T_t = T_{t-1}\cup M_t
\end{equation}
by adding the new slot entries from $M_t$ to table $T_{t-1}$. If there already exist some values for some specific slot types in $T_{t-1}$, then we will update them correspondingly using those in $M_t$ at round $t$.

The slot values in the slot filling table are combined into a formatted query by using predefined templates and then sent to the Google Flight API to fetch the results.
\section{Experiment}
We conduct two experiments to demonstrate how the new system works and evaluate its performance. The first experiment is an SLU task with one round of user feedback to correct the result if the output of the origin query is wrong. The second experiment is a multiround flight booking task with help via Amazon Mechanical Turk.
\subsection{Dataset}

We use the ATIS dataset in both experiments and follow the train/test split in \cite{liu2016attention, mesnil2015using,xu2013convolutional,wang2018bi}, which contains 4978 utterances in the training set and 893 utterances in the test set; the total number of slot tags is 127. In the second experiment, we ask Amazon Mechanical Turks to help expand the original ATIS dataset to a multiround flight booking QA dataset. For each single query in ATIS, we ask Turks to generate 1 to 4 rounds of feedback in two categories: the first type is to ask for extra information, as in Rounds 2 and 3 in Figure \ref{fig:fig1}, and the second type is to update/correct the previously stated information, as in Round 4. Round 1 uses the same feedback as that we collected for experiment 1, with one round of user feedback (to be mentioned in 3.3). The Turks are allowed to choose either type of feedback at each round by themselves (except for Round 1, which is inherited from experiment 1), and they need to note the slot tags in their feedback as ground-truth labels. The average feedback rounds include 3.2 feedback units per query.
\subsection{Model Setup}
For the RL policy model in the slot extraction model, we use Adam \cite{kingma2014adam} as the optimizer, with an initial learning rate $10^{-5}$, and we choose $\alpha = 0.5$ and $\lambda = 1$. The RNN structures used in the tagging model follow the same setup as in \cite{liu2016attention} and \cite{goo2018slot}. The nonlinear function $\phi(\cdot)$ in the reward function $R_\theta(M_t)$ is chosen as the softsign function, \emph{i.e.,} $\phi(x)=\dfrac{x}{1+\mid x \mid}$. The embedding size $k$ is set as 200.
\subsection{Experiment 1: Flight booking with one round of user feedback}
In the first experiment, we allow for our Turks to provide one round of correction feedback during both training and inference. We pretrain an RNN-based slot tagging model using the attention bi-RNN model given in \cite{liu2016attention} and the slot-gated bi-RNN model given in \cite{goo2018slot}. These models are used to generate the slot tags for the origin query and the slot candidates $C_t^{slot}$, as in Figure \ref{fig:fig2}, and are also used as baseline models. When training the new MS$_2$LU system, our Turks first check whether the template-based result generated by the slot extracted from the original user query can fetch the correct result from the Google Flight API. If the result is wrong, then our Turks will provide feedback to correct the mistake specifically. For example, if we want to query a destination ``Cincinnati", but the result displays the incorrect destination of ``Philadelphia", then a Turk should reply as follows: ``I want to go to Cincinnati actually". Similarly, we also allow Turks to have at most one round of interaction during inference for correction purposes. Table \ref{table:exp1} shows a comparison of the results by using the baseline models and those with our MS$_2$LU system. We can observe that our MS$_2$LU system with one round of feedback can improve slot F1 by more than 2.5$\%$ on both RNN tagging model structures.
\begin{table}[ht]\scriptsize
\parbox{1\linewidth}{
\centering
	\caption{Experiment 1: flight booking with one-round user feedback}
	\label{table:exp1}
	\begin{tabular}{>{\centering\arraybackslash}p{3.3cm}|>{\centering\arraybackslash}p{1.2cm}>{\centering\arraybackslash}p{1.4cm}}
		\toprule
		
		\multirow{1}{*}{\textbf{Model}} & \multirow{1}{*}{\makecell{\textbf{Slot F1 $\%$}}} &\multirow{1}{*}{\makecell{\textbf{Sentence Acc $\%$}}} \\
		\midrule
		\midrule
		
		\multirow{2}{*}{}Attention bi-RNN& 94.2 &78.9\\
		\multirow{2}{*}{}Attention bi-RNN + MS$_2$LU &\textbf{96.8} & \textbf{83.6}\\
		\midrule
		\multirow{2}{*}{}Slot-gated bi-RNN&95.2&82.6\\
		\multirow{2}{*}{}Slot-gated bi-RNN+ MS$_2$LU&\textbf{97.7}&\textbf{85.7}\\
	\bottomrule	
	\end{tabular}
}
\end{table}
\subsection{Experiment 2: Multiround flight booking task}
The second experiment is a multiround flight booking task using the expanded ATIS dataset, as described in the data section. Again, we compare the results by using the baseline RNN models and those with our MS$_2$LU system. In each round, the input to the attention RNN model is the concatenation of the raw user query and the user's feedback at round $t$. In Table \ref{table:exp2}, we show the test result by using the slot F1 scores at each round.

\begin{table}[ht]\scriptsize
\parbox{1\linewidth}{
\centering
	\caption{Experiment 2.1: Comparison of MS$_2$LU models on a multiround flight booking task}
	\label{table:exp2}
	\begin{tabular}{>{\centering\arraybackslash}p{2.4cm}|>{\centering\arraybackslash}p{0.7cm}>{\centering\arraybackslash}p{0.7cm}>{\centering\arraybackslash}p{0.7cm}>{\centering\arraybackslash}p{0.7cm}}
		\toprule
		
		\multirow{1}{*}{\textbf{Model/slot F1($\%$)}} & \multirow{1}{*}{\makecell{\textbf{Round 1}}} &\multirow{1}{*}{\makecell{\textbf{Round 2}}} &\multirow{1}{*}{\makecell{\textbf{Round 3}}} &\multirow{1}{*}{\makecell{\textbf{Round 4}}}\\
		\midrule
		\midrule
		
		\multirow{2}{*}{}Attention bi-RNN& 94.2 &93.1 &92.7&89.2\\
		\multirow{2}{*}{}Attention bi-RNN + MS$_2$LU &\textbf{96.8} & \textbf{95.3}&\textbf{95.8}&\textbf{96.3}\\
		\midrule
		\multirow{2}{*}{}Slot-gated bi-RNN&95.2&93.4&91.8&90.1\\
		\multirow{2}{*}{}Slot-gated bi-RNN+ MS$_2$LU&\textbf{97.7}&\textbf{96.5}&\textbf{96.1}&\textbf{97.1}\\
	\bottomrule	
	\end{tabular}
}
\end{table}

Based on the experimental results, we can observe that our MS$_2$LU model performs better than the baseline RNN models in all rounds. The advantage gaps of the MS$_2$LU model over baselines become larger when the number of feedback rounds increases. One main reason is that the new system is able to remember user feedback history better with an RL structure and hence can make a better decision as to whether to keep, remove or update the slot values.

Furthermore, we also compare the best performing slot-gated bi-RNN+MS$_2$LU model with two other state-of-the-art DST models, MA-DST \cite{kumar2020ma} and TRADE \cite{wu2019transferable}, to test how these DST systems perform in our multiround SLU task. The results are shown in Table \ref{table:exp2}.

\begin{table}[ht]\scriptsize
\parbox{1\linewidth}{
\centering
	\caption{Experiment 2.2: Comparison between the MS$_2$LU model and DST models on a multiround flight booking task}
	\label{table:exp2}
	\begin{tabular}{>{\centering\arraybackslash}p{2.4cm}|>{\centering\arraybackslash}p{0.7cm}>{\centering\arraybackslash}p{0.7cm}>{\centering\arraybackslash}p{0.7cm}>{\centering\arraybackslash}p{0.7cm}}
		\toprule
		
		\multirow{1}{*}{\textbf{Model/slot F1($\%$)}} & \multirow{1}{*}{\makecell{\textbf{Round 1}}} &\multirow{1}{*}{\makecell{\textbf{Round 2}}} &\multirow{1}{*}{\makecell{\textbf{Round 3}}} &\multirow{1}{*}{\makecell{\textbf{Round 4}}}\\
		\midrule
		\midrule
		
		\multirow{2}{*}{}TRADE& 83.6 &82.2 &82.5&79.8\\
		\multirow{2}{*}{}MA-DST&85.5&83.2&81.6&80.3\\
		\multirow{2}{*}{}Slot-gated bi-RNN+ MS$_2$LU&\textbf{97.7}&\textbf{96.5}&\textbf{96.1}&\textbf{97.1}\\
	\bottomrule	
	\end{tabular}
}
\end{table}
It can be observed that even the state-of-the-art DST models do not perform very well on the multiround SLU task, the main reasons for which are as follows:\\
1. We do not have any text in the system's response during training, but most of the DST systems require system response texts as one of their inputs.\\
2. Furthermore, there is no domain changing or topic changing in our multiround SLU problem, so the DST models cannot exhibit their advantages in handling the domain/topic changing scenarios as in other DST tasks.

\section{Conclusions}
In this paper, we introduce a novel multi-step SLU system that can leverage the multiround user's feedback to update slot values in a semantic frame parsing task. We test our model with two experiments on the ATIS dataset by using single-round and multiround feedback from users. By comparing with baseline tagging models, we show that our MS$_2$LU system can greatly improve the tagging model's performance by leveraging user feedback, and the advantage becomes greater when the number of feedback rounds increases.

\bibliographystyle{IEEEtran}
\bibliography{ijcai18MM}

\end{document}